\documentclass[11pt,a4paper]{article}
\usepackage[nohyperref]{emnlp2018}

\usepackage[utf8]{inputenc}
\usepackage[OT1]{fontenc}
\usepackage[english]{babel}

\usepackage{times}

\usepackage{latexsym}
\usepackage{amsmath}
\usepackage{amssymb}
\usepackage{amsfonts}

\usepackage[marginpar]{todo}
\usepackage{graphicx}
\usepackage{tikz}
\usepackage{url}

\aclfinalcopy 

\title{Using Monolingual Data in Neural Machine Translation:\newline a Systematic Study}

\author{Franck Burlot \\ %
  Lingua Custodia \\
  1, Place Charles de Gaulle \\
  78180 Montigny-le-Bretonneux \\
  {\tt\small franck.burlot@linguacustodia.com} \\\And
  François Yvon \\ %
  LIMSI, CNRS, Université Paris Saclay \\
  Campus Universitaire d'Orsay \\
  F-91~403 Orsay Cédex \\
  {\tt\footnotesize francois.yvon@limsi.fr} \\}

\begin{document}

\maketitle

\begin{abstract}
  Neural Machine Translation (MT) has radically changed the way systems are developed. A
  major difference with the previous generation (Phrase-Based MT) is the way monolingual
  target data, which often abounds, is used in these two paradigms. While Phrase-Based MT
  can seamlessly integrate very large language models trained on billions of sentences, the
  best option for Neural MT developers seems to be the generation of artificial parallel
  data through \textsl{back-translation} - a technique that fails to fully take advantage
  of existing datasets.
  In this paper, we conduct a systematic study of back-translation, comparing alternative
  uses of monolingual data, as well as multiple data generation procedures. Our findings
  confirm that back-translation is very effective and give new explanations as to why
  this is the case. We also introduce new data simulation techniques that are almost as
  effective, yet much cheaper to implement.
\end{abstract}

\section{Introduction \label{sec:introduction}}

The new generation of Neural Machine Translation (NMT) systems is known to be extremely
data hungry \cite{Koehn17sixchallenges}. Yet, most existing NMT training pipelines fail to
fully take advantage of the very large volume of monolingual source and/or parallel data
that is often available. Making a better use of data is particularly critical in domain
adaptation scenarios, where parallel adaptation data is usually assumed to be small in
comparison to out-of-domain parallel data, or to in-domain monolingual texts. This
situation sharply contrasts with the previous generation of statistical MT engines
\citep{Koehn10smt}, which could seamlessly integrate very large amounts of non-parallel 
documents, usually with a large positive effect on translation quality.

Such observations have been made repeatedly and have led to many innovative techniques to
integrate monolingual data in NMT, that we review shortly. The most successful approach to
date is the proposal of \citet{Sennrich16improving}, who use monolingual target texts to
generate artificial parallel data via backward translation (BT). This technique has since
proven effective in many subsequent studies.
It is however
very computationally costly, typically requiring to translate large sets of data. Determining 
the ``right'' amount (and quality) of BT data is another open issue, but we observe that
experiments reported in the literature only use a subset of the available monolingual 
resources. This suggests that standard recipes for BT might be sub-optimal.

This paper aims to better understand the strengths and weaknesses of
BT and to design more principled techniques to improve its effects. More
specifically, we seek to answer the following questions: since there are many ways to generate
pseudo parallel corpora, how important is the quality of this data for MT performance? Which
properties of back-translated sentences actually matter for MT quality? Does
BT act as some kind of regularizer \cite{Domhan17usingtarget}?
Can BT be efficiently simulated? Does BT data play the
same role as a target-side language modeling, or are they complementary? BT is often
used for domain adaptation: can the effect of having more in-domain data be sorted out
from the mere increase of training material \cite{Sennrich16improving}? 
For studies related to the impact of varying the size of BT data, we refer the readers to the recent work of \citet{Poncelas18investigating}.


To answer these questions, we have reimplemented several strategies to use monolingual
data in NMT 
and have run experiments on two
language pairs in a very controlled setting (see \textsection~\ref{sec:setup}). Our main
results (see \textsection~\ref{sec:stupid} and \textsection~\ref{sec:gans})
suggest promising directions for efficient domain adaptation with cheaper techniques
than conventional BT.

\section{Experimental Setup \label{sec:setup}}

\subsection{In-domain and out-of-domain data \label{ssec:data}}



We are mostly interested with the following training scenario: a large out-of-domain parallel corpus, and limited monolingual in-domain data. We 
focus here on the \emph{Europarl} domain, for which we have ample
 data in several languages, and use as in-domain training data the
 Europarl corpus\footnote{Version~7, see \url{www.statmt.org/europarl/}.} \cite{Koehn05europarl} for two translation directions:
English$\rightarrow$German and English$\rightarrow$French. As we study the benefits of monolingual data, most of our experiments only use the target side of this corpus. The rationale for choosing this domain is to  (i)
to perform large scale comparisons of synthetic and natural parallel corpora; (ii) to study the
effect of BT in a well-defined domain-adaptation scenario. 
For both language pairs, we use the Europarl tests from 2007 and 2008\footnote{\url{www.statmt.org/wmt08}.}
for evaluation purposes, keeping test 2006 for development. When measuring out-of-domain
performance, we will use the WMT newstest 2014.


\subsection{NMT setups and performance \label{ssec:baseline-setup}}

Our baseline NMT system implements the attentional encoder-decoder
approach \cite{Cho2014gru,Bahdanau15neural} as implemented in Nematus
\cite{nematus17} on 4 million out-of-domain parallel sentences.
For French we use samples from News-Commentary-11 and Wikipedia from WMT 2014 shared translation task,
as well as the Multi-UN \cite{multiun-corpus} and EU-Bookshop \cite{SkadinsEA:LREC14} corpora.
For German, we use samples from News-Commentary-11, Rapid, Common-Crawl (WMT 2017) and Multi-UN
(see table~\ref{tab:data}).
Bilingual BPE units \cite{Sennrich16BPE} are learned with 50k merge operations, yielding
vocabularies of about respectively 32k and 36k for English$\rightarrow$French and 32k and 
44k for English$\rightarrow$German.

\begin{table}[t]
\begin{center} \small
\begin{tabular}{lcc|cc}
\hline
& \multicolumn{2}{c}{Out-of-domain} & \multicolumn{2}{c}{In-domain} \\
& Sents & Token & Sents & Token \\
\hline
\textbf{en-fr} & 4.0M & 86.8M/97.8M & 1.9M & 46.0M/50.6M \\
\textbf{en-de} & 4.1M & 84.5M/77.8M & 1.8M & 45.5M/43.4M \\
\hline
\end{tabular}
\end{center}
  \caption{Size of parallel corpora}
  \label{tab:data}
\end{table}

Both systems use 512-dimensional word embeddings and a single hidden layer with 1024 cells. They are
optimized using Adam \cite{Kingma2014adam} and early stopped according to the validation
performance. Training lasted for about three weeks on an Nvidia K80 GPU card.

Systems generating back-translated data are trained using the same out-of-domain corpus,
where we simply exchange the source and target sides. They are further documented in 
\textsection~\ref{ssec:quality}. 

For the sake of comparison, we also train a system that has access to a large batch of in-domain
parallel data following the strategy often referred to as ``fine-tuning'':
upon convergence of the baseline model, we resume training
with a 2M sentence in-domain corpus mixed with an equal amount of randomly selected
out-of-domain natural sentences, with the same architecture and training parameters,
running validation every 2000 updates with a patience of~10.
Since BPE units are selected based only on the out-of-domain
statistics, fine-tuning is performed on sentences that are slightly longer (ie.\ they
contain more units) than for the initial training. This system defines an upper-bound of the
translation performance and is denoted below as \texttt{natural}.

%
Our baseline and topline results are in Table~\ref{tab:bt-quality}, where we
measure translation performance using BLEU \cite{Papineni:2002:acl},
BEER \cite{Stanojevic14beer} (higher is better) and characTER \cite{Wang16character}
(smaller is better). As they are trained from much smaller amounts of data than current
systems, these baselines are not quite competitive to today's best system, but still represent
serious baselines for these datasets. Given our setups, fine-tuning with in-domain natural data improves 
BLEU by almost 4 points for both translation directions on in-domain tests; it also improves, albeit by a 
smaller margin, the BLEU score of the out-of-domain tests.



\section{Using artificial parallel data in NMT \label{sec:artificial}}

A simple way to use monolingual data in MT is to turn it into synthetic parallel data and let
the training procedure run as usual \cite{Bojar11improvingtranslation}. In this
section, we explore various ways to implement this strategy. We first reproduce results of
\citet{Sennrich16improving} with BT of various qualities, that we then analyze thoroughly. 

\begin{table*}[t]
\begin{center} \small
\begin{tabular}{lccc | ccc | ccc}
\hline
\multicolumn{10}{c}{\textbf{English$\rightarrow$French}}\\
\hline
& \multicolumn{3}{c}{\textbf{test-07}} & \multicolumn{3}{c}{\textbf{test-08}} & \multicolumn{3}{c}{\textbf{newstest-14}} \\
& BLEU & BEER & CTER & BLEU & BEER & CTER & BLEU & BEER & CTER \\
\texttt{Baseline}                  & 31.25 & 62.14 & 51.89 & 32.17 & 62.35 & 50.79 & 33.06 & 61.97 & 48.56 \\
\texttt{backtrans-bad}    & 31.55 & 62.39 & 51.50 & 31.89 & 62.23 & 51.73 & 31.99 & 61.59 & 48.86 \\
\texttt{backtrans-good}   & 32.99 & 63.43 & 49.58 & 33.25 & 63.08 & 49.29 & 33.52 & 62.62 & 47.23 \\
\texttt{backtrans-nmt}    & 33.30 & 63.33 & 50.02 & 33.39 & 63.09 & 49.48 & 34.11 & 62.76 & 46.94 \\
\texttt{fwdtrans-nmt}     & 31.93 & 62.55 & 50.84 & 32.62 & 62.66 & 49.83 & 33.56 & 62.44 & 47.65 \\
\texttt{backfwdtrans-nmt} & 33.09 & 63.19 & 50.08 & 33.70 & 63.25 & 48.83 & 34.00 & 62.76 & 47.22 \\
\texttt{natural}          & 35.10 & 64.71 & 48.33 & 35.29 & 64.52 & 48.26 & 34.96 & 63.08 & 46.67 \\
\hline
\multicolumn{10}{c}{\textbf{English$\rightarrow$German}} \\
\hline
& \multicolumn{3}{c}{\textbf{test-07}} & \multicolumn{3}{c}{\textbf{test-08}} & \multicolumn{3}{c}{\textbf{newstest-14}} \\
& BLEU & BEER & CTER & BLEU & BEER & CTER & BLEU & BEER & CTER \\
\texttt{Baseline}         & 21.36 & 57.08 & 63.32 & 21.27 & 57.11 & 60.67 & 22.49 & 57.79 & 55.64 \\
\texttt{backtrans-bad}    & 21.84 & 57.85 & 61.24 & 21.04 & 57.44 & 59.77 & 22.28 & 57.70 & 55.49 \\
\texttt{backtrans-good}   & 23.33 & 59.03 & 58.84 & 23.11 & 57.14 & 57.14 & 22.87 & 58.09 & 54.91 \\
\texttt{backtrans-nmt}    & 23.00 & 59.12 & 58.31 & 23.10 & 58.85 & 56.67 & 22.91 & 58.12 & 54.67 \\
\texttt{fwdtrans-nmt}     & 21.97 & 57.46 & 61.99 & 21.89 & 57.53 & 59.71 & 22.52 & 57.93 & 55.13 \\
\texttt{backfwdtrans-nmt} & 22.99 & 58.37 & 60.45 & 22.82 & 58.14 & 58.80 & 23.04 & 58.17 & 54.96 \\
\texttt{natural}          & 26.74 & 61.14 & 56.19 & 26.16 & 60.64 & 54.76 & 23.84 & 58.64 & 54.23 \\
\hline
\end{tabular}
\end{center}
  \caption{Performance {\it wrt.} different BT qualities}
  \label{tab:bt-quality}
\end{table*}

\begin{table*}[t]
\begin{center} \small
\begin{tabular}{lcccc|cccc}
\hline
& \multicolumn{4}{c}{\textbf{French$\rightarrow$English}} & \multicolumn{4}{c}{\textbf{German$\rightarrow$English}} \\
  \hline
 & \textbf{test-07} & \textbf{test-08} & \textbf{nt-14} & \textbf{unk} & \textbf{test-07} & \textbf{test-08} & \textbf{nt-14} & \textbf{unk} \\
\texttt{backtrans-bad}  & 18.86 & 19.27 & 20.49 & 3.22\% & 14.66 & 14.62 & 15.07 & 1.45\% \\
\texttt{backtrans-good} & 29.71 & 29.51 & 32.10 & 0.24\% & 24.19 & 24.19 & 25.75 & 0.73\% \\
\texttt{backtrans-nmt}  & 31.10 & 31.43 & 31.27 & 0.0\%  & 26.02 & 26.03 & 26.98 & 0.0\% \\
\hline
\end{tabular}
\end{center}
  \caption{BLEU scores for (backward) translation into English}
  \label{tab:back}
\end{table*}

\subsection{The quality of Back-Translation \label{ssec:quality}}

\subsubsection{Setups}
BT requires the availability of an MT system in the reverse translation direction. We consider here
three MT systems of increasing quality:
\begin{enumerate}
\item \texttt{backtrans-bad}: this is a very poor SMT system trained using only 50k parallel
  sentences from the out-of-domain data, and no additional monolingual data. For
  this system as for the next one, we use Moses \cite{Koehn07moses} out-of-the-box, computing
  alignments with Fastalign \cite{Dyer13simple}, with a minimal pre-processing (basic tokenization). 
  This setting provides us with a
  pessimistic estimate of what we could get in low-resource conditions.
\item \texttt{backtrans-good}: these are much larger SMT systems, which use the same
  parallel data as the baseline NMTs (see \textsection~\ref{ssec:baseline-setup}) and all the
  English monolingual data available for the WMT~2017 shared tasks, totalling
  approximately 174M sentences. These systems are strong, yet relatively cheap
  to build.
\item \texttt{backtrans-nmt}: these are the best NMT systems we could train, using
  settings that replicate the forward translation NMTs.
\end{enumerate}

\begin{figure*}[tb]
\begin{center}
\begin{tabular}{cc}
\hspace{-9mm} \includegraphics[scale=0.37]{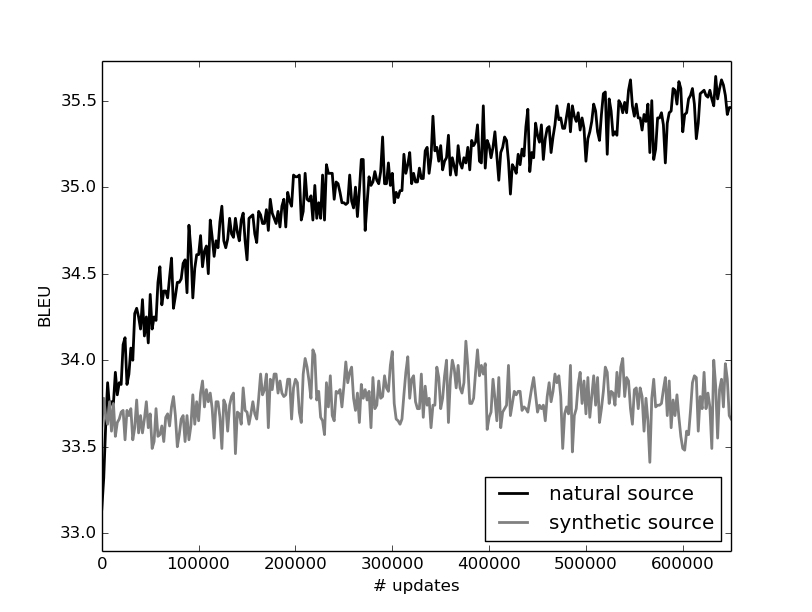} \hspace{-6mm} & \hspace{-6mm} \includegraphics[scale=0.37]{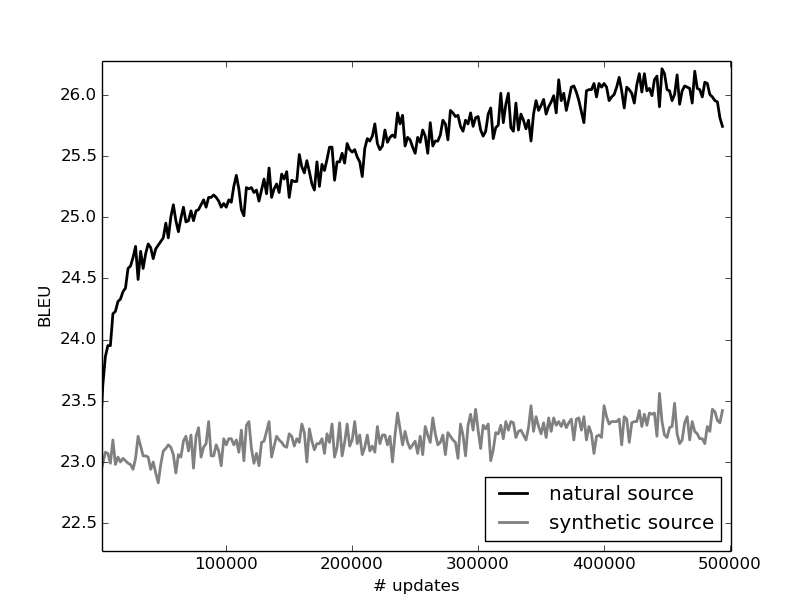} \\
English$\rightarrow$French & English$\rightarrow$German \\
\end{tabular}
\end{center}
\vspace{-4mm}
\label{fig:overfit}
\caption{Learning curves from \texttt{backtrans-nmt} and \texttt{natural}.
  Artificial parallel data is more prone to overfitting than natural data.}
\end{figure*}

Note that we do not use any in-domain (\emph{Europarl}) data to train these systems.
Their performance is reported in Table~\ref{tab:back}, where we observe
a 12 BLEU points gap between the worst and best systems (for both languages).

As noted \textsl{eg.}\ in \cite{Park17building,Crego16neuralmachine},
artificial parallel data obtained through \emph{forward-translation} (FT) can also prove
advantageous and we also consider a FT system (\texttt{fwdtrans-nmt}): in this case the
\emph{target} side of the corpus is artificial and is generated using the baseline NMT applied to
a natural source.

\subsubsection{BT quality does matter}

Our results (see Table~\ref{tab:bt-quality}) replicate the findings of \cite{Sennrich16improving}: large gains can be
obtained from BT (nearly $+2$~BLEU in French and German); better artificial data
yields better translation systems. Interestingly, our best Moses system is almost as good
as the NMT and an order of magnitude faster to train. Improvements obtained with the bad system
are much smaller; contrary to the better MTs, this system is even detrimental for the out-of-domain test.

Gains with forward translation are significant, as in \cite{ChineaRios17adapting}, albeit
about half as good as with BT, and result in small improvements for the in-domain
and for the out-of-domain tests. Experiments combining forward and backward translation
(\texttt{backfwdtrans-nmt}), each using a half of the available artificial data, do not outperform
the best BT results.

We finally note the large remaining difference between BT data and natural data, even though they 
only differ in their source side. This shows that at least in our domain-adaptation settings, BT
does not really act as a regularizer, contrarily to the findings of
\cite{Poncelas18investigating,Sennrich16BPE}. Figure~\ref{fig:overfit} displays the learning curves
of these two systems. We observe that \texttt{backtrans-nmt} improves quickly
in the earliest updates and then stays horizontal, whereas \texttt{natural}
continues improving, even after 400k updates. Therefore BT
does not help to avoid overfitting, it actually encourages it, which may be due ``easier''
training examples (cf.\ \textsection~\ref{ssec:BT-properties}).


\subsection{Properties of back-translated data \label{ssec:BT-properties}}
\begin{figure*}[t!]
\begin{center}
\begin{tabular}{cc}
\hspace{-9mm} \includegraphics[scale=0.37]{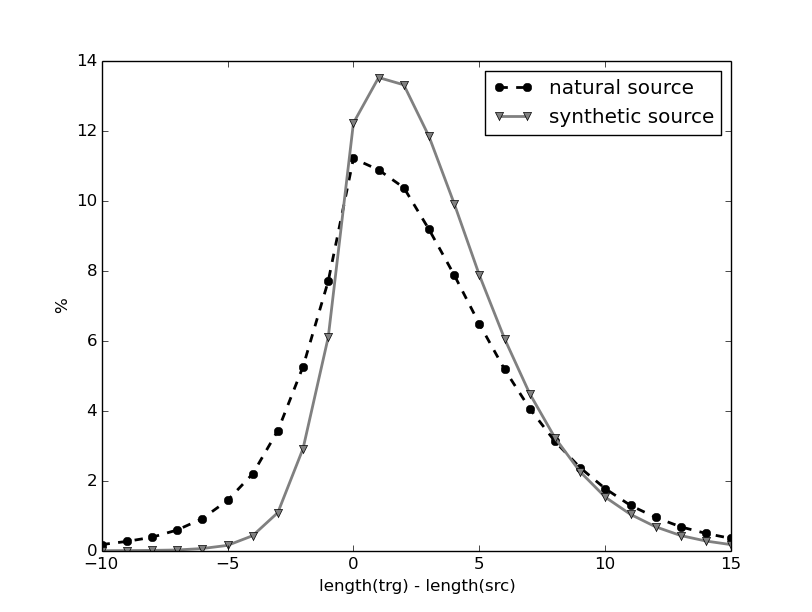} \hspace{-6mm} & \hspace{-6mm} \includegraphics[scale=0.37]{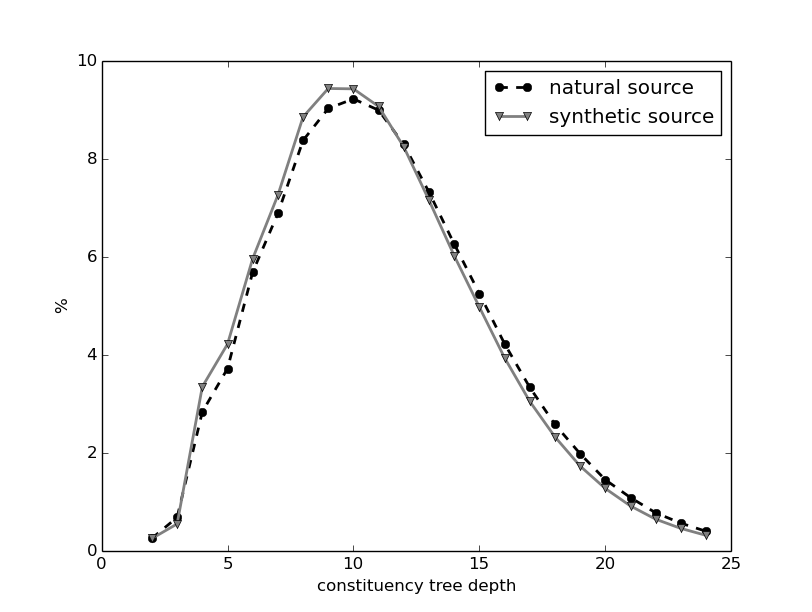} \\
(a) & (b) \\
\hspace{-6mm} \includegraphics[scale=0.37]{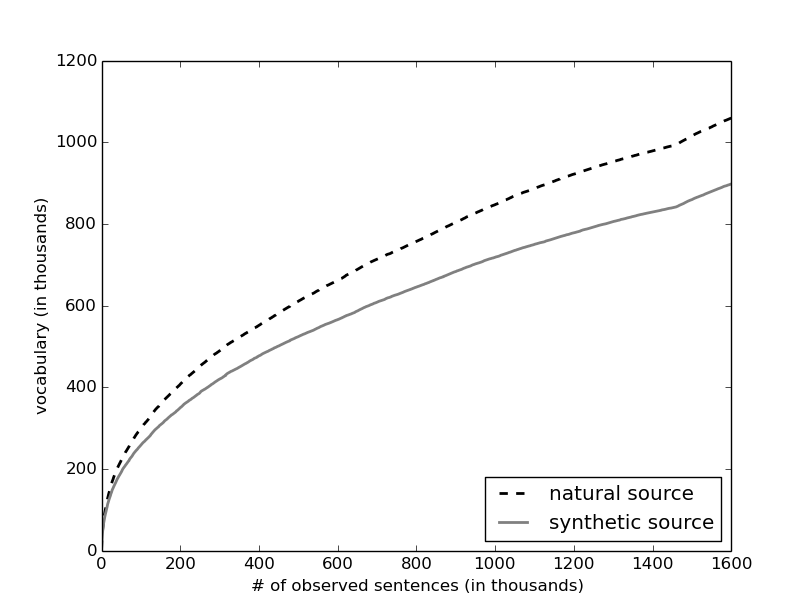} \hspace{-6mm} & \hspace{-6mm} \includegraphics[scale=0.37]{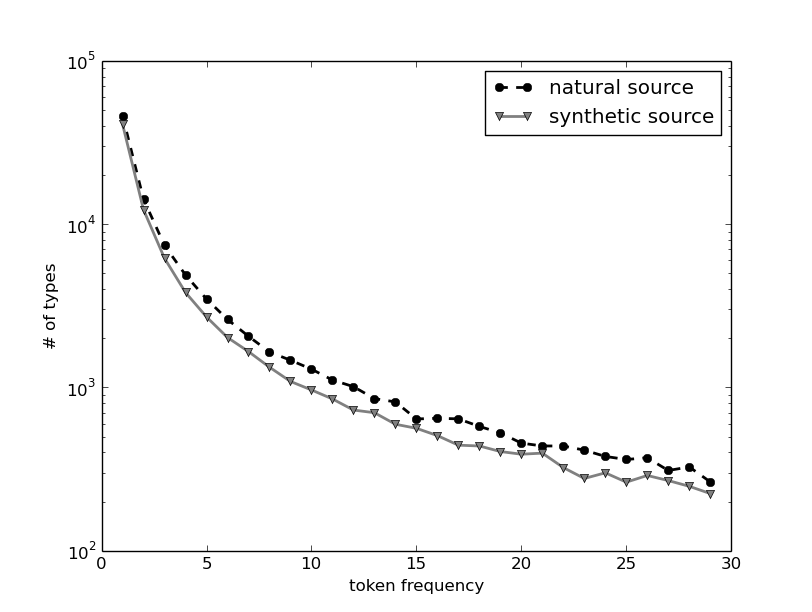} \\
(c) & (d) \\
\end{tabular}
\end{center}
\vspace{-4mm}
\caption{Properties of pseudo-English data obtained with \texttt{backtrans-nmt} from French.
The synthetic source contains shorter sentences (a) and slightly simpler syntax (b).
The vocabulary growth {\it wrt.} an increasing number of observed sentences (c)
and the token-type correlation (d) suggest that the natural source is lexically richer.
}
\label{fig:bt-properties-fr}
\end{figure*}

\begin{figure*}[t] 
\begin{center}
\begin{tabular}{cc}
\hspace{-9mm} \includegraphics[scale=0.37]{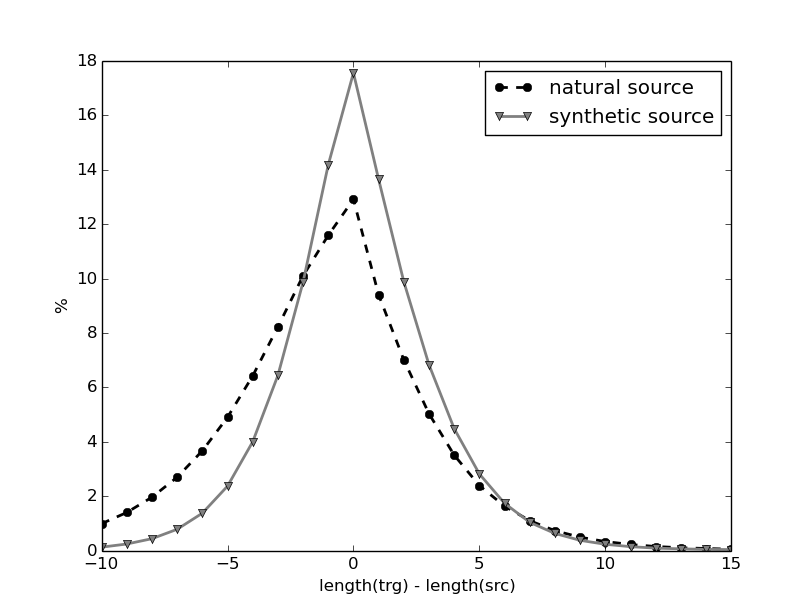} \hspace{-6mm} & \hspace{-6mm} \includegraphics[scale=0.37]{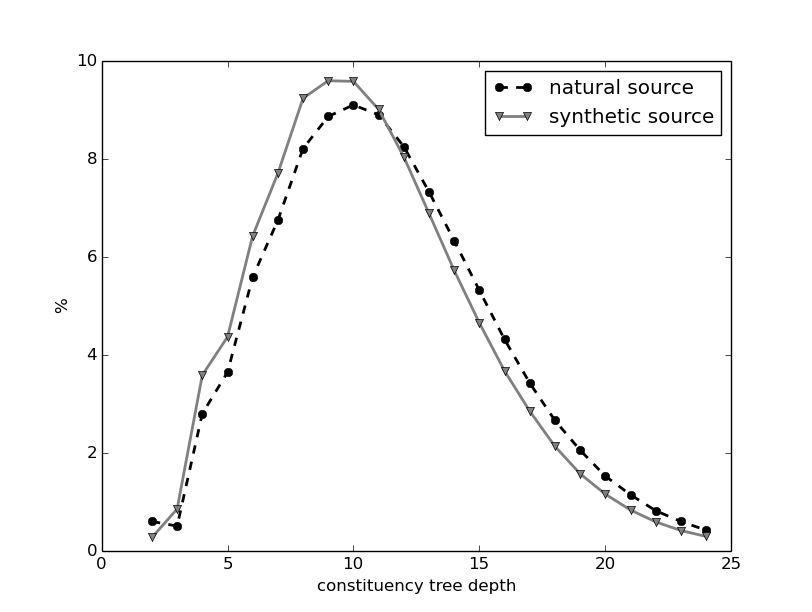} \\
(a) & (b) \\
\hspace{-6mm} \includegraphics[scale=0.37]{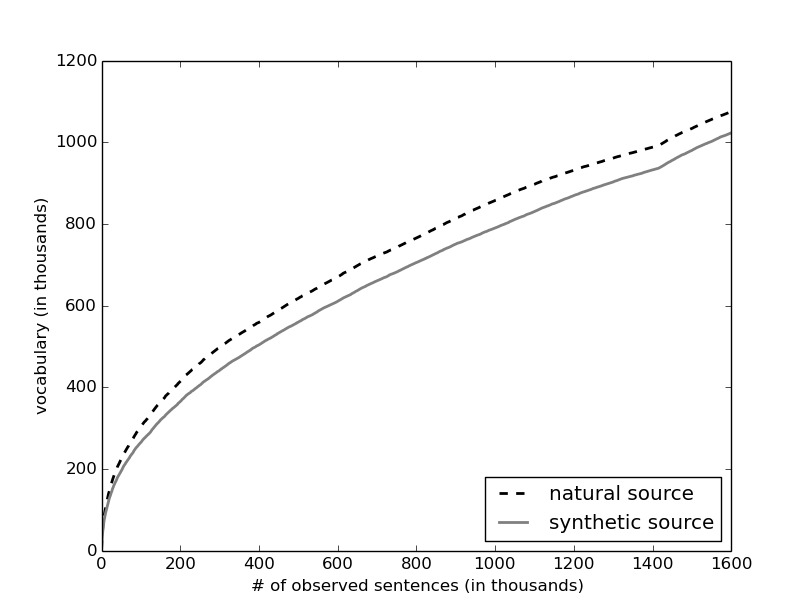} \hspace{-6mm} & \hspace{-6mm} \includegraphics[scale=0.37]{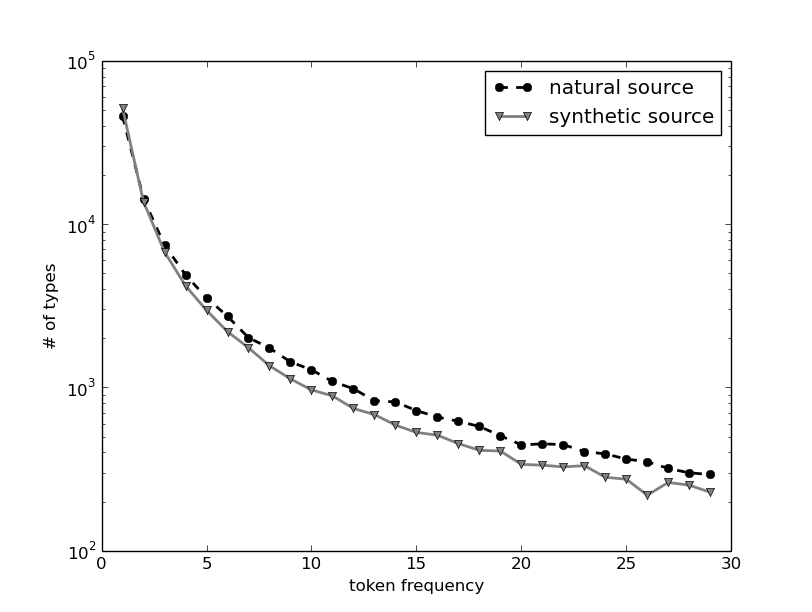} \\
(c) & (d) \\
\end{tabular}
\end{center}
\vspace{-4mm}
\caption{Properties of pseudo-English data obtained with \texttt{backtrans-nmt} (back-translated from German).
Tendencies similar to English-French can be observed and difference in syntax complexity is even more visible.
}
\label{fig:bt-properties-de}
\end{figure*}

Comparing the natural and artificial sources of our parallel data {\it wrt.} several
linguistic and distributional properties, we observe that (see Fig.~\ref{fig:bt-properties-fr} - \ref{fig:bt-properties-de}):
\begin{itemize}
\item[(i)] artificial sources are on average shorter than natural ones: when using BT, cases where the source is
  shorter than the target are rarer; cases when they have the same length are more frequent.
\item[(ii)] automatic word alignments between artificial sources tend to be more
  monotonic than when using natural sources, as measured by the average Kendall $\tau$
  of source-target alignments \cite{Birch10lrscore}: for French-English the respective
  numbers are 0.048 (natural) and 0.018 (artificial); for German-English 0.068 and
  0.053. Using more monotonic sentence pairs turns out to be a facilitating factor for NMT, 
  as also noted by \citet{Crego16neuralmachine}.
\item[(iii)] 
  syntactically, artificial sources are simpler than real data;
  We observe significant differences in the distributions of tree depths.\footnote{Parses were automatically computed with CoreNLP \cite{corenlp14}.}
 \item[(iv)] distributionally, plain word occurrences in artificial sources are more concentrated; this also translates into
   both a slower increase of the number of types {\it wrt.} the number of sentences and a smaller 
   number of rare events. 
\end{itemize}

The intuition is that properties (i) and (ii) should help translation as compared to natural
source, while property (iv) should be detrimental. We checked (ii) by building systems with only 10M
words from the natural parallel data selecting these data either randomly or based on the regularity of their
word alignments. Results in Table~\ref{tab:selection} show that the latter is much preferable for the 
overall performance. This might explain that the mostly monotonic BT from Moses are 
almost as good as the fluid BT from NMT and that both boost the baseline.


\begin{table*}[t] \small
\begin{center}
\begin{tabular}{lccc | ccc | ccc}
\hline
& \multicolumn{3}{c}{\textbf{test-07}} & \multicolumn{3}{c}{\textbf{test-08}} & \multicolumn{3}{c}{\textbf{newstest-14}} \\
& BLEU & BEER & CTER & BLEU & BEER & CTER & BLEU & BEER & CTER \\
random     & 32.08 & 62.98 & 50.78 & 32.66 & 62.86 & 49.99 & 23.05 & 55.38 & 58.51 \\
monotonic & 33.52 & 63.75 & 49.51 & 33.73 & 63.59 & 48.91 & 32.16 & 61.75 & 48.64 \\
\hline
\end{tabular}
\end{center}
\caption{Selection strategies for BT data (English-French) \label{tab:selection}}
\end{table*}

\section{Stupid Back-Translation \label{sec:stupid}}

We now analyze the effect of using much simpler data generation schemes,
which do not require the availability of a backward translation engine. 
\subsection{Setups}
We use the following cheap ways to generate pseudo-source texts:
\begin{enumerate}
\item \texttt{copy}: in this setting, the source side is a mere copy of the target-side
  data. Since the source vocabulary of the
  NMT is fixed, copying the target sentences
  can cause the occurrence of OOVs.
  To avoid this situation, \citet{Currey17copied} decompose
  the target words into source-side units to make the copy look like
  source sentences. Each OOV found in the copy is split
  into smaller units until all the resulting chunks are
  in the source vocabulary.
\item \texttt{copy-marked}: another way to integrate copies without having to deal with
  OOVs is to augment the source vocabulary with a copy of the target vocabulary. In this
  setup, \citet{Ha2016} ensure that both vocabularies never overlap by marking the target
  word copies with a special language identifier. Therefore the English word {\it resume}
  cannot be confused with the homographic French word, which is marked {\it @fr@resume}.
\item \texttt{copy-dummies}: instead of using actual copies, we replace each word with
  ``dummy'' tokens. We use this unrealistic setup to observe the training over noisy
  and hardly informative source sentences.
\end{enumerate}

We then use the procedures described in~\textsection~\ref{ssec:baseline-setup}, except that the pseudo-source
embeddings in the \texttt{copy-marked} setup are pretrained for
three epochs on the in-domain data, while all remaining parameters are frozen. This prevents 
random parameters from hurting the already trained model.

\subsection{Copy+marking+noise is not so stupid}

We observe that the \texttt{copy} setup has only a small
impact on the English-French system, for which the
baseline is already strong. This is less true for
English-German where simple copies yield a significant 
improvement. Performance drops for both language pairs in the
\texttt{copy-dummies} setup.

\begin{table*}[thb]
\begin{center} \small
\begin{tabular}{lccc | ccc | ccc}
\hline
\multicolumn{10}{c}{\textbf{English$\rightarrow$French}}\\
\hline
& \multicolumn{3}{c}{\textbf{test-07}} & \multicolumn{3}{c}{\textbf{test-08}} & \multicolumn{3}{c}{\textbf{newstest-14}} \\
& BLEU & BEER & CTER & BLEU & BEER & CTER & BLEU & BEER & CTER \\
Baseline              & 31.25 & 62.14 & 51.89 & 32.17 & 62.35 & 50.79 & 33.06 & 61.97 & 48.56 \\
\texttt{copy}         & 31.65 & 62.45 & 52.09 & 32.23 & 62.37 & 52.20 & 32.80 & 61.99 & 49.05 \\
\texttt{copy-dummies} & 30.89 & 62.06 & 52.07 & 31.51 & 61.98 & 51.46 & 31.43 & 60.92 & 50.58 \\
\texttt{copy-marked}  & 32.01 & 62.66 & 51.57 & 32.31 & 62.52 & 51.46 & 32.33 & 61.55 & 49.44 \\
+ Source noise        & 31.87 & 62.52 & 52.69 & 32.64 & 62.55 & 51.63 & 33.04 & 62.11 & 48.47 \\
\hline
\multicolumn{10}{c}{\textbf{English$\rightarrow$German}} \\
\hline
& \multicolumn{3}{c}{\textbf{test-07}} & \multicolumn{3}{c}{\textbf{test-08}} & \multicolumn{3}{c}{\textbf{newstest-14}} \\
& BLEU & BEER & CTER & BLEU & BEER & CTER & BLEU & BEER & CTER \\
Baseline              & 21.36 & 57.08 & 63.32 & 21.27 & 57.11 & 60.67 & 22.49 & 57.79 & 55.64 \\
\texttt{copy}         & 22.15 & 57.95 & 61.49 & 21.95 & 57.72 & 59.58 & 22.59 & 57.83 & 55.44 \\
\texttt{copy-dummies} & 21.73 & 57.84 & 61.35 & 21.38 & 57.38 & 60.10 & 21.12 & 56.81 & 57.21 \\
\texttt{copy-marked}  & 22.58 & 58.23 & 61.10 & 22.47 & 57.97 & 59.24 & 22.53 & 57.54 & 55.85 \\
+ Source noise        & 22.92 & 58.62 & 60.27 & 22.83 & 58.36 & 58.48 & 22.34 & 57.47 & 55.72 \\
\hline
\end{tabular}
\end{center}
  \caption{Performance {\it wrt.} various stupid BTs}
  \label{tab:bt-stupid}
\end{table*}

We achieve our best gains with the \texttt{copy-marked} setup, which is the best way to use a copy of the target (although
the performance on the out-of-domain tests is at most the same as the baseline).
Such gains may look surprising, since the NMT model does not need to learn to translate but only
to copy the source. This is indeed what happens: to confirm
this, we built a fake test set having identical source and target side (in French). The average cross-entropy for this test set is 0.33, very close to 0, to be compared with an average cost of 58.52 when we process an actual source (in English). This means that the model
has learned to copy words from source to target with no difficulty, even for sentences not seen in training. 
A follow-up question is whether training a copying
task instead of a translation task limits the improvement: would the NMT learn better if the task
was harder? To measure this, we introduce noise in the target sentences copied onto the
source, following the procedure of \citet{lample17mono}: it deletes random words and 
performs a small random permutation of the remaining words. Results
({\it + Source noise}) show no difference for the French in-domain test sets, but bring the
out-of-domain score to the level of the baseline. Finally, we observe a significant
improvement on German in-domain test sets, compared to the baseline (about +1.5 BLEU).
This last setup is even almost as good as the \texttt{backtrans-nmt} condition
(see \textsection~\ref{ssec:quality}) for German. This shows that learning to reorder
and predict missing words can more effectively serve our purposes
than simply learning to copy.


\section{Towards more natural pseudo-sources \label{sec:gans}}

Integrating monolingual data into NMT can be as easy as copying the target into
the source, which already gives some improvement; adding noise makes things even better. 
We now study ways to make pseudo-sources look more like natural data, using the
framework of Generative Adversarial Networks (GANs) \cite{goodfellow14gans}, an idea
borrowed from \citet{lample17mono}\footnote{Our implementation is available at
\url{https://github.com/franckbrl/nmt-pseudo-source-discriminator}}. 


\subsection{GAN setups}

\begin{table*}[t]
\begin{center} \small
\begin{tabular}{lccc | ccc | ccc}
\hline
\multicolumn{10}{c}{\textbf{English$\rightarrow$French}}\\
\hline
& \multicolumn{3}{c}{\textbf{test-07}} & \multicolumn{3}{c}{\textbf{test-08}} & \multicolumn{3}{c}{\textbf{newstest-14}} \\
& BLEU & BEER & CTER & BLEU & BEER & CTER & BLEU & BEER & CTER \\
Baseline                     & 31.25 & 62.14 & 51.89 & 32.17 & 62.35 & \textbf{50.79} & \textbf{33.06} & 61.97 & 48.56 \\
\texttt{copy-marked}         & 32.01 & 62.66 & \textbf{51.57} & 32.31 & 62.52 & 51.46 & 32.33 & 61.55 & 49.44 \\
+ GANs                       & 31.95 & 62.55 & 52.87 & 32.24 & 62.47 & 52.16 & 32.86 & 61.90 & 48.97  \\
\texttt{copy-marked} + noise & 31.87 & 62.52 & 52.69 & 32.64 & 62.55 & 51.63 & 33.04 & 62.11 & 48.47 \\
+ GANs                       & \textbf{32.41} & \textbf{62.78} & 52.25 & \textbf{32.79} & \textbf{62.72} & 50.92 & 33.01 & \textbf{61.98} & \textbf{48.37}  \\
\hline
\texttt{backtrans-nmt}       & \textbf{33.30} & \textbf{63.33} & \textbf{50.02} & \textbf{33.39} & \textbf{63.09} & \textbf{49.48} & \textbf{34.11} & \textbf{62.76} & \textbf{46.94} \\
+ Distinct encoders          & 32.29 & 62.83 & 51.55 & 32.98 & 62.91 & 51.19 & 33.60 & 62.43 & 48.06  \\
+ GANs                       & 32.91 & 63.08 & 51.17 & 33.24 & 62.93 & 50.82 & 33.77 & 62.42 & 47.80  \\
\hline
\texttt{natural}             & 35.10 & 64.71 & 48.33 & 35.29 & 64.52 & 48.26 & 34.96 & 63.08 & 46.67 \\
\hline
\multicolumn{10}{c}{\textbf{English$\rightarrow$German}} \\
\hline
& \multicolumn{3}{c}{\textbf{test-07}} & \multicolumn{3}{c}{\textbf{test-08}} & \multicolumn{3}{c}{\textbf{newstest-14}} \\
& BLEU & BEER & CTER & BLEU & BEER & CTER & BLEU & BEER & CTER \\
Baseline                     & 21.36 & 57.08 & 63.32 & 21.27 & 57.11 & 60.67 & 22.49 & \textbf{57.79} & 55.64 \\
\texttt{copy-marked}         & 22.58 & 58.23 & 61.10 & 22.47 & 57.97 & 59.24 & 22.53 & 57.54 & 55.85 \\
+ GANs                       & 22.71 & 58.25 & 61.25 & 22.44 & 57.86 & 59.28 & \textbf{22.81} & 57.54 & 55.99  \\
\texttt{copy-marked} + noise & 22.92 & 58.62 & 60.27 & \textbf{22.83} & \textbf{58.36} & \textbf{58.48} & 22.34 & 57.47 & 55.72 \\
+ GANs                       & \textbf{23.01} & \textbf{58.66} & \textbf{60.22} & 22.53 & 58.16 & 58.65 & 22.64 & 57.70 & \textbf{55.48}  \\
\hline
\texttt{backtrans-nmt}       & 23.00 & 59.12 & \textbf{58.31} & 23.10 & 58.85 & \textbf{56.67} & 22.91 & 58.12 & \textbf{54.67} \\
+ Distinct encoders          & 23.62 & 58.83 & 59.74 & 23.10 & \textbf{58.50} & 58.19 & 22.82 & \textbf{57.91} & 54.96  \\
+ GANs                       & \textbf{23.65} & \textbf{58.85} & 59.70 & \textbf{23.20} & \textbf{58.50} & 58.22 & \textbf{23.00} & 57.89 & 55.15  \\
\hline
\texttt{natural}             & 26.74 & 61.14 & 56.19 & 26.16 & 60.64 & 54.76 & 23.84 & 58.64 & 54.23 \\
\hline
\end{tabular}
\end{center}
  \caption{Performance {\it wrt.} different GAN setups}
  \label{tab:bt-gans}
\end{table*}

In our setups, we use a marked target copy, viewed as a {\it fake} source, which a
\emph{generator} encodes so as to fool a discriminator \emph{trained} to distinguish a {\it fake} from
a {\it natural} source. Our architecture contains two distinct encoders, one for the
natural source and one for the pseudo-source. The latter acts as the generator ($G$) in
the GAN framework, computing a representation of the pseudo-source that is then input
to a discriminator ($D$), which has to sort natural from artificial encodings. 
$D$ assigns a probability of a sentence being natural.

During training, the cost of the discriminator is computed over two batches, one with
natural (out-of-domain) sentences $\textbf{x}$ and one with (in-domain) pseudo-sentences $\textbf{x}^\prime$.
The discriminator is a bidirectional-Recurrent Neural Network (RNN) of dimension~1024.
Averaged states are passed to a single feed-forward
layer, to which a sigmoid is applied. It inputs encodings of natural ($E(\textbf{x})$) and
pseudo-sentences ($G(\textbf{x}^\prime)$) and is trained to optimize:
\vspace{-5mm}
\begin{center}
\begin{align*} 
J^{(D)} = & - \frac{1}{2} \mathbb{E}_{\textbf{x} \sim p_{\text{real}}} \log D(E(\textbf{x})) \\
        & - \frac{1}{2} \mathbb{E}_{\textbf{x}^\prime \sim p_{\text{pseudo}}} \log (1 - D(G(\textbf{x}^\prime)))
\end{align*}
\end{center}

$G$'s parameters are updated to maximally fool $D$, thus the loss $J^{(G)}$:
\[
J^{(G)} = - \mathbb{E}_{\textbf{x}^\prime \sim p_{\text{pseudo}}} \log D(G(\textbf{x}^\prime))
\]

Finally, we keep the usual MT objective.  ($\textbf{s}$ is a real or pseudo-sentence):
\[
J^{(\text{MT})} = \log p(\textbf{y} | \textbf{s}) =  - \mathbb{E}_{\textbf{s} \sim p_{\text{all}}} \log \text{MT}(\textbf{s})
\]
We thus need to train three sets of parameters: $ \theta^{(D)}, \theta^{(G)} $ and $\theta^{(\text{MT})}$ (MT parameters), with
$\theta^{(G)} \in \theta^{(\text{MT})}$. 
The pseudo-source encoder and embeddings are updated {\it wrt.} both $J^{(G)}$ and $J^{(\text{MT})}$.
Following \cite{Goyal16professorforcing}, $\theta^{(G)}$ is updated only when $D$'s
accuracy exceeds $75\%$. On the other hand, $\theta^{(D)}$ is not updated
when its accuracy exceeds $99\%$. At each update, two batches are generated for each
type of data, which are encoded with the real or pseudo-encoder.  The encoder outputs
serve to compute $J^{(D)}$ and $J^{(G)}$. Finally, the pseudo-source is
encoded again (once $G$ is updated), both encoders are plugged into the translation model
and the MT cost is back-propagated down to the real and pseudo-word embeddings.
Pseudo-encoder and discriminator parameters are pre-trained for 10k updates.
At test time, the pseudo-encoder is ignored and inference is run as usual.

\begin{table*}
\begin{center} \small
\begin{tabular}{lccc | ccc | ccc}
\hline
\multicolumn{10}{c}{\textbf{English$\rightarrow$French}}\\
\hline
& \multicolumn{3}{c}{\textbf{test-07}} & \multicolumn{3}{c}{\textbf{test-08}} & \multicolumn{3}{c}{\textbf{newstest-14}} \\
& BLEU & BEER & CTER & BLEU & BEER & CTER & BLEU & BEER & CTER \\
Baseline           & 31.25 & 62.14 &  \textbf{51.89} & 32.17 & 62.35 & \textbf{50.79} & 33.06 & 61.97 & 48.56 \\
\texttt{deep-fusion} & 31.85 & 62.52 & 52.27 & 32.25 & 62.40 & 51.64 & \textbf{33.65} & \textbf{62.40} & \textbf{48.24} \\
\texttt{copy-marked} + noise + GANs & \textbf{32.41} & \textbf{62.78} & 52.25 & \textbf{32.79} & \textbf{62.72} & 50.92 & 33.01 & 61.98 & 48.37  \\
+\texttt{deep-fusion} & 31.96 & 62.59 & 51.96 & 32.59 & 62.59 & 51.65 & 32.96 & 61.95 & 48.95 \\
\hline
\multicolumn{10}{c}{\textbf{English$\rightarrow$German}} \\
\hline
& \multicolumn{3}{c}{\textbf{test-07}} & \multicolumn{3}{c}{\textbf{test-08}} & \multicolumn{3}{c}{\textbf{newstest-14}} \\
& BLEU & BEER & CTER & BLEU & BEER & CTER & BLEU & BEER & CTER \\
Baseline           & 21.36 & 57.08 & 63.32 & 21.27 & 57.11 & 60.67 & 22.49 & 57.79 & 55.64 \\
\texttt{deep-fusion} & 21.65 & 57.57 & 62.38 & 21.33 & 57.33 & 60.54 & \textbf{23.10} & \textbf{58.06} & \textbf{55.33} \\
\texttt{copy-marked} + noise + GANs & \textbf{23.01} & \textbf{58.66} & \textbf{60.22} & \textbf{22.53} & \textbf{58.16} & \textbf{58.65} & 22.64 & 57.70 & 55.48  \\
+\texttt{deep-fusion} & 23.07 & 58.50 & 60.47 & 22.86 & 58.18 & 58.76 & 22.64 & 57.46 & 55.85 \\
\hline
\end{tabular}
\end{center}
  \caption{Deep-fusion model}
  \label{tab:deep-fusion}
\end{table*}

\subsection{GANs can help}

Results are in Table~\ref{tab:bt-gans}, assuming the same fine-tuning procedure as above.
On top of the \texttt{copy-marked} setup, our GANs
do not provide any improvement in both language
pairs, with the exception of a small improvement for English-French
on the out-of-domain test, which we understand as a sign
that the model is more robust to domain variations, just like
when adding pseudo-source noise. When combined with noise,
the French model yields the best performance we could
obtain with stupid BT on the in-domain tests, at least in terms of BLEU and BEER.
On the News domain, we remain close to the baseline level,
with slight improvements in German.

A first observation is that this method brings stupid BT models closer to conventional BT, at a greatly reduced computational cost.
While French still remains 0.4 to 1.0 BLEU below very good backtranslation, both approaches are in the same ballpark for German
- may be because BTs are better for the former system than for the latter.

Finally note that the GAN architecture has two differences with basic \texttt{copy-marked}: (a) a distinct encoder for real and pseudo-sentence; (b) a different training regime for these encoders. To sort out the effects of (a) and (b), we reproduce the GAN setup with BT sentences, instead of copies. Using a separate encoder for the pseudo-source in the \texttt{backtrans-nmt} setup can be detrimental to performance  (see Table~\ref{tab:bt-gans}): translation degrades in French for all metrics.
Adding GANs on top of the pseudo-encoder was not able to make up for the degradation observed in French, but allowed the German system to slightly
outperform \texttt{backtrans-nmt}.
Even though this setup is unrealistic and overly costly, it shows that GANs are actually helping even good systems.

\section{Using Target Language Models \label{sec:lms}}

In this section, we compare the previous methods with the use of a target side Language Model (LM).
Several proposals exist in the literature to integrate LMs in NMT:
for instance, \citet{Domhan17usingtarget} strengthen the decoder by integrating an extra, source
independent, RNN layer in a conventional NMT architecture. 
Training is performed either with parallel, or monolingual data. In the latter case, word prediction only relies on the
source independent part of the network. 

\subsection{LM Setup}
We have followed \citet{Gulcehre17onintegrating} and
reimplemented\footnote{Our implementation is part of the Nematus toolkit (theano branch):
\url{https://github.com/EdinburghNLP/nematus/blob/theano/doc/deep_fusion_lm.md}}
their \texttt{deep-fusion} technique. It
requires to first independently learn a RNN-LM on the in-domain target
data with a cross-entropy objective; then to train the optimal combination 
of the translation and the language models by adding
the hidden state of the RNN-LM as an additional input to the softmax layer of the
decoder. 

Our RNN-LMs are trained using dl4mt\footnote{\url{https://github.com/nyu-dl/dl4mt-tutorial}}
with the target side of the parallel data and
the Europarl corpus (about 6M sentences for both French and German),
using a one-layer GRU with the same dimension as the MT decoder
(1024). 

\subsection{LM Results}

Results are in Table~\ref{tab:deep-fusion}.
They show that \texttt{deep-fusion} hardly improves
the Europarl results, while we obtain about +0.6 BLEU 
over the baseline on newstest-2014 for both languages.
\texttt{deep-fusion} differs from stupid BT in that the model
is not directly optimized on the in-domain data, but uses the
LM trained on Europarl to maximize the likelihood
of the out-of-domain training data. Therefore, no specific
improvement is to be expected in terms of domain adaptation,
and the performance increases in the more general domain.
Combining \texttt{deep-fusion} and \texttt{copy-marked} + noise + GANs
brings slight improvements on the German in-domain test sets,
and performance out of the domain remains near the baseline level.


\section{Re-analyzing the effects of BT \label{sec:freezing}}

\begin{table*}
\begin{center} \small
\begin{tabular}{lccc|ccc}
\hline
& \multicolumn{3}{c}{\textbf{English$\rightarrow$French}} & \multicolumn{3}{c}{\textbf{English$\rightarrow$German}} \\
  \hline
 & \textbf{test-07} & \textbf{test-08} & \textbf{nt-14} & \textbf{test-07} & \textbf{test-08} & \textbf{nt-14} \\
Baseline                & 31.25 & 32.17 & 33.06 & 21.36 & 21.27 & 22.49 \\
\hline
\texttt{backtrans-nmt}  & 33.30 & 33.39 & 34.11 & 23.00 & 23.10 & 22.91 \\
+ Freeze source embedd. & 33.20 & 33.24 & 34.16 & 22.84 & 22.85 & 23.00 \\
+ Freeze encoder        & 33.17 & 33.25 & 33.73 & 22.72 & 22.74 & 22.64 \\
+ Freeze attention      & 33.13 & 33.22 & 33.47 & 23.03 & 23.01 & 22.85 \\
\hline
\texttt{copy-marked}  & 32.01 & 32.31 & 32.33 & 22.58 & 22.47 & 22.53 \\
+ Freeze encoder        & 31.70 & 32.39 & 32.90 & 22.59 & 22.30 & 22.81 \\
+ Freeze attention      & 31.59 & 32.39 & 32.54 & 22.55 & 22.13 & 22.69 \\
\hline
\texttt{fwdtrans-nmt}   & 31.93 & 32.62 & 33.56 & 21.97 & 21.89 & 22.52 \\
+ Freeze decoder        & 31.84 & 32.62 & 33.35 & 21.91 & 21.65 & 13.61 \\
\hline
\texttt{natural}        & 35.10 & 35.29 & 34.96 & 26.74 & 26.16 & 23.84 \\
+ Freeze encoder        & 34.02 & 34.25 & 34.09 & 24.95 & 25.08 & 23.44 \\
+ Freeze attention      & 34.13 & 34.42 & 34.19 & 25.13 & 24.97 & 23.35 \\
\hline
\end{tabular}
\end{center}
\caption{BLEU scores with selective parameter freezing}
\label{tab:freezing}
\end{table*}

As a follow up of previous discussions, we analyze the effect of BT on the internals of the network.
Arguably, using a copy of the target sentence instead of a natural source should not be helpful
for the encoder, but is it also the case with a strong BT? What are the effects on the attention model?

\subsection{Parameter freezing protocol}

To investigate these questions, we run the same fine-tuning using the \texttt{copy-marked}, 
\texttt{backtrans-nmt} and \texttt{backtrans-nmt} setups. Note that except for the last one, 
all training scenarios have access to same target training data. We intend
to see whether the overall performance of the NMT system degrades when we
selectively freeze certain sets of parameters, meaning that they are not updated during fine-tuning.

\subsection{Results}
BLEU scores are in Table~\ref{tab:freezing}. 
The \texttt{backtrans-nmt} setup is hardly impacted by selective updates: updating
the only decoder leads to a degradation of at most 0.2 BLEU.  
For \texttt{copy-marked}, we were not able to freeze the source embeddings, since these are initialized when fine-tuning begins and
therefore need to be trained. We observe that freezing the encoder and/or the attention
parameters has no impact on the English-German system, whereas it slightly degrades the
English-French one.  This suggests that using artificial sources, even of the poorest quality, has a
positive impact on all the components of the network, which makes another big difference
with the LM integration scenario. 

The largest degradation is for \texttt{natural}, where the model is prevented from learning from
informative source sentences, which leads to a decrease of 0.4 to over 1.0 BLEU. We
assume from these experiments that BT impacts most of all the decoder, and learning to
encode a pseudo-source, be it a copy or an actual back-translation, only marginally helps to
significantly improve the quality. Finally, in the \texttt{fwdtrans-nmt} setup,
freezing the decoder does not seem to harm learning with a natural source.


\section{Related work}

The literature devoted to the use of monolingual data is large, and quickly expanding. We already alluded to several possible ways to use such data: using back- or forward-translation or using a target language model. The former approach is mostly documented in \cite{Sennrich16improving}, and recently analyzed in \cite{Park17building}, which focus on fully artificial settings as well as pivot-based artificial data; and \cite{Poncelas18investigating}, which studies the effects of increasing the size of BT data. The studies of \citet{Crego16neuralmachine,Park17building} also consider forward translation and \citet{ChineaRios17adapting} expand these results to domain adaptation scenarios. Our results are complementary to these earlier studies.
 
As shown above, many alternatives to BT exist. The most obvious is to use target LMs \cite{Domhan17usingtarget,Gulcehre17onintegrating}, as we have also done here; but attempts to improve the encoder using multi-task learning also exist \cite{Zhang16exploiting}.

This investigation is also related to recent attempts to consider supplementary data with a valid
target side, such as multi-lingual NMT \cite{Firat16multiway}, where source
texts in several languages are fed in the same encoder-decoder architecture, with
partial sharing of the layers. This is another realistic scenario where additional resources can be used 
to selectively improve parts of the model.

Round trip training is another important source of inspiration, as it can be viewed as a way to use BT to perform  semi-unsupervised \cite{Yong16semisupervised} or unsupervised \cite{He16dualleaning} training of NMT. The most convincing attempt to date along these lines has been proposed by \citet{lample17mono}, who propose to use GANs to mitigate the difference between artificial and natural data.

\section{Conclusion \label{sec:conclusion}}

In this paper, we have analyzed various ways
to integrate monolingual data in an NMT framework, focusing
on their impact on quality and domain adaptation. While confirming the effectiveness of BT, our study also
proposed significantly cheaper ways to improve the baseline performance,
using a slightly modified copy of the target, instead of its full BT.
When no high quality BT is available, using GANs to
make the pseudo-source sentences closer to natural source sentences
is an efficient solution for domain adaptation.

To recap our answers to our initial questions: the quality of BT actually matters for NMT (cf.\ \textsection~\ref{ssec:quality}) and it seems that, even though artificial source are lexically less diverse and syntactically complex than real sentence, their monotonicity is a facilitating factor. We have studied cheaper alternatives and found out that copies of the target, if properly noised (\textsection~\ref{sec:stupid}), and even better, if used with GANs, could be almost as good as low quality BTs (\textsection~\ref{sec:gans}): BT is only worth its cost when good BT can be generated. Finally, BT seems preferable to integrating external LM - at least in our data condition (\textsection~\ref{sec:lms}). Further experiments with larger LMs are needed to confirm this observation, and also to evaluate the complementarity of both strategies. More work is needed to better understand the impact of BT on subparts of the network (\textsection~\ref{sec:freezing}). 

In future work, we plan to investigate other cheap ways to generate artificial data. The experimental setup we proposed
may also benefit from a refining of the data selection strategies to focus on the most useful monolingual sentences. 



\bibliographystyle{acl_natbib_nourl}
\bibliography{monolingual-paper}

\begin{thebibliography}{34}
\expandafter\ifx\csname natexlab\endcsname\relax\def\natexlab#1{#1}\fi

\bibitem[{Bahdanau et~al.(2015)Bahdanau, Cho, and Bengio}]{Bahdanau15neural}
Dzmitry Bahdanau, Kyunghyun Cho, and Yoshua Bengio. 2015.
\newblock Neural machine translation by jointly learning to align and
  translate.
\newblock In \emph{Proceedings of the first International Conference on
  Learning Representations}, San Diego, CA.

\bibitem[{Birch and Osborne(2010)}]{Birch10lrscore}
Alexandra Birch and Miles Osborne. 2010.
\newblock {LRscore} for evaluating lexical and reordering quality in {MT}.
\newblock In \emph{Proceedings of the Joint Fifth Workshop on Statistical
  Machine Translation and MetricsMATR}, WMT '10, pages 327--332, Stroudsburg,
  PA, USA. Association for Computational Linguistics.

\bibitem[{Bojar and Tamchyna(2011)}]{Bojar11improvingtranslation}
Ond\v{r}ej Bojar and Ale\v{s} Tamchyna. 2011.
\newblock Improving translation model by monolingual data.
\newblock In \emph{Proceedings of the Sixth Workshop on Statistical Machine
  Translation}, WMT '11, pages 330--336, Stroudsburg, PA, USA. Association for
  Computational Linguistics.

\bibitem[{Cheng et~al.(2016)Cheng, Xu, He, He, Wu, Sun, and
  Liu}]{Yong16semisupervised}
Yong Cheng, Wei Xu, Zhongjun He, Wei He, Hua Wu, Maosong Sun, and Yang Liu.
  2016.
\newblock Semi-supervised learning for neural machine translation.
\newblock In \emph{Proceedings of the 54th Annual Meeting of the Association
  for Computational Linguistics (Volume 1: Long Papers)}, pages 1965--1974.
  Association for Computational Linguistics.

\bibitem[{Chinea-Rios et~al.(2017)Chinea-Rios, Álvaro Peris, and
  Casacuberta}]{ChineaRios17adapting}
Mara Chinea-Rios, Álvaro Peris, and Francisco Casacuberta. 2017.
\newblock Adapting neural machine translation with parallel synthetic data.
\newblock In \emph{Proceedings of the Second Conference on Machine Translation,
  Volume 1: Research Papers}, pages 138--147, Copenhagen, Denmark. Association
  for Computational Linguistics.

\bibitem[{Cho et~al.(2014)Cho, van Merrienboer, Bahdanau, and
  Bengio}]{Cho2014gru}
Kyunghyun Cho, Bart van Merrienboer, Dzmitry Bahdanau, and Yoshua Bengio. 2014.
\newblock On the properties of neural machine translation: Encoder-decoder
  approaches.
\newblock In \emph{Proceedings of SSST-8, Eighth Workshop on Syntax, Semantics
  and Structure in Statistical Translation}, pages 103--111, Doha, Qatar.
  Association for Computational Linguistics.

\bibitem[{Crego and Senellart(2016)}]{Crego16neuralmachine}
Josep~Maria Crego and Jean Senellart. 2016.
\newblock Neural machine translation from simplified translations.
\newblock \emph{CoRR}, abs/1612.06139.

\bibitem[{Currey et~al.(2017)Currey, Miceli~Barone, and
  Heafield}]{Currey17copied}
Annad Currey, Antonio~Valerio Miceli~Barone, and Kenneth Heafield. 2017.
\newblock Copied monolingual data improves low-resource neural machine
  translation.
\newblock In \emph{Proceedings of the Second Conference on Machine
  Translation}, pages 148--156, Copenhagen, Denmark. Association for
  Computational Linguistics.

\bibitem[{Domhan and Hieber(2017)}]{Domhan17usingtarget}
Tobias Domhan and Felix Hieber. 2017.
\newblock Using target-side monolingual data for neural machine translation
  through multi-task learning.
\newblock In \emph{Proceedings of the 2017 Conference on Empirical Methods in
  Natural Language Processing}, pages 1500--1505, Copenhagen, Denmark.
  Association for Computational Linguistics.

\bibitem[{Dyer et~al.(2013)Dyer, Chahuneau, and Smith}]{Dyer13simple}
Chris Dyer, Victor Chahuneau, and Noah~A. Smith. 2013.
\newblock {A Simple, Fast, and Effective Reparameterization of IBM Model 2}.
\newblock In \emph{Proceedings of the 2013 Conference of the North American
  Chapter of the Association for Computational Linguistics: Human Language
  Technologies}, pages 644--648, Atlanta, Georgia.

\bibitem[{Eisele and Chen(2010)}]{multiun-corpus}
Andreas Eisele and Yu~Chen. 2010.
\newblock {MultiUN: A Multilingual Corpus from United Nation Documents}.
\newblock In \emph{Proceedings of the Seventh conference on International
  Language Resources and Evaluation}, pages 2868--2872. European Language
  Resources Association (ELRA).

\bibitem[{Firat et~al.(2016)Firat, Cho, and Bengio}]{Firat16multiway}
Orhan Firat, Kyunghyun Cho, and Yoshua Bengio. 2016.
\newblock Multi-way, multilingual neural machine translation with a shared
  attention mechanism.
\newblock In \emph{Proceedings of the 2016 Conference of the North American
  Chapter of the Association for Computational Linguistics: Human Language
  Technologies}, pages 866--875. Association for Computational Linguistics.

\bibitem[{Goodfellow et~al.(2014)Goodfellow, Pouget-Abadie, Mirza, Xu,
  Warde-Farley, Ozair, Courville, and Bengio}]{goodfellow14gans}
Ian Goodfellow, Jean Pouget-Abadie, Mehdi Mirza, Bing Xu, David Warde-Farley,
  Sherjil Ozair, Aaron Courville, and Yoshua Bengio. 2014.
\newblock Generative adversarial nets.
\newblock In Z.~Ghahramani, M.~Welling, C.~Cortes, N.~D. Lawrence, and K.~Q.
  Weinberger, editors, \emph{Advances in Neural Information Processing Systems
  27}, pages 2672--2680. Curran Associates, Inc.

\bibitem[{Goyal et~al.(2016)Goyal, Lamb, Zhang, Zhang, Courville, and
  Bengio}]{Goyal16professorforcing}
Anirudh Goyal, Alex Lamb, Ying Zhang, Saizheng Zhang, Aaron~C. Courville, and
  Yoshua Bengio. 2016.
\newblock Professor forcing: {A} new algorithm for training recurrent networks.
\newblock In \emph{Advances in Neural Information Processing Systems 29: Annual
  Conference on Neural Information Processing Systems 2016}, pages 4601--4609,
  Barcelona, Spain.

\bibitem[{Gulcehre et~al.(2017)Gulcehre, Firat, Xu, Cho, and
  Bengio}]{Gulcehre17onintegrating}
Caglar Gulcehre, Orhan Firat, Kelvin Xu, Kyunghyun Cho, and Yoshua Bengio.
  2017.
\newblock On integrating a language model into neural machine translation.
\newblock \emph{Comput. Speech Lang.}, 45(C):137--148.

\bibitem[{Ha et~al.(2016)Ha, Niehues, and Waibel}]{Ha2016}
Thanh-Le Ha, Jan Niehues, and Alex Waibel. 2016.
\newblock Toward multilingual neural machine translation with universal encoder
  and decoder.
\newblock In \emph{Proceedings of the 13th International Workshop on Spoken
  Language Translation (IWSLT 2016)}, Seattle, WA, USA.

\bibitem[{He et~al.(2016)He, Xia, Qin, Wang, Yu, Liu, and Ma}]{He16dualleaning}
Di~He, Yingce Xia, Tao Qin, Liwei Wang, Nenghai Yu, Tieyan Liu, and Wei-Ying
  Ma. 2016.
\newblock Dual learning for machine translation.
\newblock In D.~D. Lee, M.~Sugiyama, U.~V. Luxburg, I.~Guyon, and R.~Garnett,
  editors, \emph{Advances in Neural Information Processing Systems 29}, pages
  820--828. Curran Associates, Inc.

\bibitem[{Kingma and Ba(2014)}]{Kingma2014adam}
Diederik Kingma and Jimmy Ba. 2014.
\newblock Adam: A method for stochastic optimization.
\newblock \emph{arXiv preprint arXiv:1412.6980}.

\bibitem[{Koehn(2005)}]{Koehn05europarl}
Philipp Koehn. 2005.
\newblock A parallel corpus for statistical machine translation.
\newblock In \emph{Proc. MT-Summit}, Phuket, Thailand.

\bibitem[{Koehn(2010)}]{Koehn10smt}
Philipp Koehn. 2010.
\newblock \emph{Statistical Machine Translation}.
\newblock Cambridge University Press.

\bibitem[{Koehn et~al.(2007)Koehn, Hoang, Birch, Callison-Burch, Federico,
  Bertoldi, Cowan, Shen, Moran, Zens, Dyer, Bojar, Constantin, and
  Herbst}]{Koehn07moses}
Philipp Koehn, Hieu Hoang, Alexandra Birch, Chris Callison-Burch, Marcello
  Federico, Nicola Bertoldi, Brooke Cowan, Wade Shen, Christine Moran, Richard
  Zens, Chris Dyer, Ond{\v{r}}ej Bojar, Alexandra Constantin, and Evan Herbst.
  2007.
\newblock Moses: Open source toolkit for statistical {MT}.
\newblock In \emph{Proc. ACL:Systems Demos}, pages 177--180, Prague, Czech
  Republic.

\bibitem[{Koehn and Knowles(2017)}]{Koehn17sixchallenges}
Philipp Koehn and Rebecca Knowles. 2017.
\newblock Six challenges for neural machine translation.
\newblock In \emph{Proceedings of the First Workshop on Neural Machine
  Translation}, pages 28--39, Vancouver. Association for Computational
  Linguistics.

\bibitem[{Lample et~al.(2017)Lample, Denoyer, and Ranzato}]{lample17mono}
Guillaume Lample, Ludovic Denoyer, and Marc'Aurelio Ranzato. 2017.
\newblock Unsupervised machine translation using monolingual corpora only.
\newblock \emph{CoRR}, abs/1711.00043.

\bibitem[{Manning et~al.(2014)Manning, Surdeanu, Bauer, Finkel, Bethard, and
  McClosky}]{corenlp14}
Christopher Manning, Mihai Surdeanu, John Bauer, Jenny Finkel, Steven Bethard,
  and David McClosky. 2014.
\newblock {The Stanford CoreNLP Natural Language Processing Toolkit}.
\newblock In \emph{Proceedings of 52nd Annual Meeting of the Association for
  Computational Linguistics: System Demonstrations}, pages 55--60, Baltimore,
  Maryland. Association for Computational Linguistics.

\bibitem[{Papineni et~al.(2002)Papineni, Roukos, Ward, and
  Zhu}]{Papineni:2002:acl}
Kishore Papineni, Salim Roukos, Todd Ward, and Wei-Jing Zhu. 2002.
\newblock Bleu: a method for automatic evaluation of machine translation.
\newblock In \emph{Proceedings of the 40th Annual Meeting on Association for
  Computational Linguistics}, ACL '02, pages 311--318, Stroudsburg, PA, USA.

\bibitem[{Park et~al.(2017)Park, Song, and Yoon}]{Park17building}
Jaehong Park, Jongyoon Song, and Sungroh Yoon. 2017.
\newblock Building a neural machine translation system using only synthetic
  parallel data.
\newblock \emph{CoRR}, abs/1704.00253.

\bibitem[{Poncelas et~al.(2018)Poncelas, Shterionov, Way, de~Buy~Wenniger, and
  Passban}]{Poncelas18investigating}
Alberto Poncelas, Dimitar Shterionov, Andy Way, Gideon~Maillette
  de~Buy~Wenniger, and Peyman Passban. 2018.
\newblock Investigating backtranslation in neural machine translation.
\newblock In \emph{Proceedings of the 21st Annual Conference of the European
  Association for Machine Translation}, EAMT, Alicante, Spain.

\bibitem[{Sennrich et~al.(2017)Sennrich, Firat, Cho, Birch, Haddow, Hitschler,
  Junczys-Dowmunt, L\"{a}ubli, Miceli~Barone, Mokry, and Nadejde}]{nematus17}
Rico Sennrich, Orhan Firat, Kyunghyun Cho, Alexandra Birch, Barry Haddow,
  Julian Hitschler, Marcin Junczys-Dowmunt, Samuel L\"{a}ubli, Antonio~Valerio
  Miceli~Barone, Jozef Mokry, and Maria Nadejde. 2017.
\newblock Nematus: a toolkit for neural machine translation.
\newblock In \emph{Proceedings of the Software Demonstrations of the 15th
  Conference of the European Chapter of the Association for Computational
  Linguistics}, pages 65--68, Valencia, Spain. Association for Computational
  Linguistics.

\bibitem[{Sennrich et~al.(2016{\natexlab{a}})Sennrich, Haddow, and
  Birch}]{Sennrich16improving}
Rico Sennrich, Barry Haddow, and Alexandra Birch. 2016{\natexlab{a}}.
\newblock Improving neural machine translation models with monolingual data.
\newblock In \emph{Proceedings of the 54th Annual Meeting of the Association
  for Computational Linguistics (Volume 1: Long Papers)}, pages 86--96, Berlin,
  Germany. Association for Computational Linguistics.

\bibitem[{Sennrich et~al.(2016{\natexlab{b}})Sennrich, Haddow, and
  Birch}]{Sennrich16BPE}
Rico Sennrich, Barry Haddow, and Alexandra Birch. 2016{\natexlab{b}}.
\newblock Neural machine translation of rare words with subword units.
\newblock In \emph{Proceedings of the 54th Annual Meeting of the Association
  for Computational Linguistics (Volume 1: Long Papers)}, pages 1715--1725.
  Association for Computational Linguistics.

\bibitem[{Skadi\c{n}\v{s} et~al.(2014)Skadi\c{n}\v{s}, Tiedemann, Rozis, and
  Deksne}]{SkadinsEA:LREC14}
Raivis Skadi\c{n}\v{s}, J{\"o}rg Tiedemann, Roberts Rozis, and Daiga Deksne.
  2014.
\newblock Billions of parallel words for free: Building and using the eu
  bookshop corpus.
\newblock In \emph{Proceedings of the 9th International Conference on Language
  Resources and Evaluation (LREC-2014)}, Reykjavik, Iceland. European Language
  Resources Association (ELRA).

\bibitem[{Stanojevi\'{c} and Sima'an(2014)}]{Stanojevic14beer}
Milo\v{s} Stanojevi\'{c} and Khalil Sima'an. 2014.
\newblock Fitting sentence level translation evaluation with many dense
  features.
\newblock In \emph{Proceedings of the 2014 Conference on Empirical Methods in
  Natural Language Processing (EMNLP)}, pages 202--206, Doha, Qatar.
  Association for Computational Linguistics.

\bibitem[{Wang et~al.(2016)Wang, Peter, Rosendahl, and Ney}]{Wang16character}
Weiyue Wang, Jan-Thorsten Peter, Hendrik Rosendahl, and Hermann Ney. 2016.
\newblock {C}harac{T}er: {T}ranslation {E}dit {R}ate on {C}haracter {L}evel.
\newblock In \emph{Proceedings of the First Conference on Machine Translation},
  pages 505--510, Berlin, Germany. Association for Computational Linguistics.

\bibitem[{Zhang and Zong(2016)}]{Zhang16exploiting}
Jiajun Zhang and Chengqing Zong. 2016.
\newblock Exploiting source-side monolingual data in neural machine
  translation.
\newblock In \emph{Proceedings of the 2016 Conference on Empirical Methods in
  Natural Language Processing}, pages 1535--1545, Austin, Texas. Association
  for Computational Linguistics.

\end{thebibliography}

\end{document}